\documentclass[conference]{IEEEtran}
\IEEEoverridecommandlockouts
\usepackage{cite}
\usepackage{amsmath,amssymb,amsfonts}
\usepackage{algorithmic}
\usepackage{graphicx}
\usepackage{textcomp}
\usepackage{nicefrac}
\usepackage{xcolor}\usepackage[ruled,vlined,algo2e,linesnumbered]{algorithm2e}
\usepackage{hyperref}
\makeatletter
\def\UrlAlphabet{%
      \do\a\do\b\do\c\do\d\do\e\do\f\do\g\do\h\do\i\do\j%
      \do\k\do\l\do\m\do\n\do\o\do\p\do\q\do\r\do\s\do\t%
      \do\u\do\v\do\w\do\x\do\y\do\z\do\A\do\B\do\C\do\D%
      \do\E\do\F\do\G\do\H\do\I\do\J\do\K\do\L\do\M\do\N%
      \do\O\do\P\do\Q\do\R\do\S\do\T\do\U\do\V\do\W\do\X%
      \do\Y\do\Z}
\def\UrlDigits{\do\1\do\2\do\3\do\4\do\5\do\6\do\7\do\8\do\9\do\0}
\g@addto@macro{\UrlBreaks}{\UrlOrds}
\g@addto@macro{\UrlBreaks}{\UrlAlphabet}
\g@addto@macro{\UrlBreaks}{\UrlDigits}
\makeatother
\def\BibTeX{{\rm B\kern-.05em{\sc i\kern-.025em b}\kern-.08em
    T\kern-.1667em\lower.7ex\hbox{E}\kern-.125emX}}
\begin{document}

\title{GPNAS: A Neural Network Architecture Search Framework Based on Graphical Predictor}

\author{\IEEEauthorblockN{1\textsuperscript{st} Dige Ai}
\IEEEauthorblockA{\textit{dept. Renmin University of China} \\
Beijing, China \\
aidg@stu.cpu.edu.com}
\and
\IEEEauthorblockN{2\textsuperscript{nd} Hong Zhang}
\IEEEauthorblockA{\textit{dept. Massachusetts Institute of Technology} \\
Massachusetts, USA \\
lishanzai@gmail.com}
}

\maketitle

\begin{abstract}
In practice, the problems encountered in Neural Architecture Search (NAS) training are not simple problems, but often a series of difficult combinations (wrong compensation estimation, curse of dimension, overfitting, high complexity, etc.). In this paper, we propose a framework to decouple network structure from operator search space, and use two BOHBs to search alternatively. Considering that activation function and initialization are also important parts of neural network, the generalization ability of the model will be affected. We introduce an activation function and an initialization method domain, and add them into the operator search space to form a generalized search space, so as to improve the generalization ability of the child model. We then trained a GCN-based predictor using feedback from the child model. This can not only improve the search efficiency, but also solve the problem of dimension curse. Next, unlike other NAS studies, we used predictors to analyze the stability of different network structures. Finally, we applied our framework to neural structure search and achieved significant improvements on multiple datasets.
\end{abstract}


\section{Introduction}
Due to the amazing characterization ability of deep neural networks (DNNs), it has achieved very good performance in various tasks, such as target detection \cite{Ren2015Faster,Liu2016SSD,Redmon2016You}, natural language processing \cite{Gehring2017Convolutional,Vaswani2017Attention,Devlin2018BERT}, speech recognition \cite{Graves2013Speech,Oord2016WaveNet}, face recognition \cite{deng2018arcface,fr2,fr3}, etc. Early neural network architecture searches were manual, using a multi-experiment approach to configure a good network by optimizing and adjusting different operators and their internal parameters.

In recent studies, NAS is considered to be a statistics-based model that can be optimized. Its goal is to find an optimal architecture that satisfies certain constraints, the most important of which is the search space. In general, the search space is coupled by a set of candidates for operator and skip-connection. Operators are used to implement linear and nonlinear transformations, or even splices, during data flow, such as max-pooling and 5*5 convolution operations. Skip-connections describe how operators are connected within a network. They transform an otherwise linear network structure into a complex topology.
Based on the above search space, many valuable works have been proposed by researchers to find better network architectures. Some studies have adopted evolutionary algorithms \cite{liu2018progressive,van2018evolutionary,real2018regularized,liu2017hierarchical}, which require a large amount of computation. Later, reinforcement learning methods (RL-based methods)\cite{pham2018efficient,cai2018path,cai2018proxylessnas} and gradient-based methods \cite{liu2018darts,cai2018proxylessnas,xie2018snas,Wu2018FBNet} were designed, which reduced the calculation cost, but failed to solve the problem of credit allocation. Zela\cite{zela_towards_2018} used BOHB to implement the search of neural network architecture, and they combined operators in a reasonable way to form a compound function. In fact, this is a kind of manual prior, which requires the engineer to have strong model design ability, which we believe defeates the original purpose of NAS design. Liu\cite{liu2018darts} proposed a differentiable neural network architecture search model, which can directly optimize the searcher through gradient descent method. However, the searched cell structure will not be directly applied to the network required by the engineering. Instead, the cells in the tiny network will be transplanted to the larger network structure through an agent mechanism. In this way, although the search speed is improved, the training goal and evaluation goal of the searcher are not unified, which may lead to overfitting.

Jiang\cite{jiang_neural_2019} believed that the search paths of skip-connection and operator are different, and the search space of the two should not be coupled.Therefore, they decomposed the search space of skip-connection and operator, and designed an alternate search model based on RL, in which two controllers were used to search in the search space of both.However, their analysis of the search path of skip-connection has a large noise, and we believe that it is inefficient to search for skip-connection using RL. In order to catalyze search, Liu\cite{liu2018progressive} designed a predictor that can predict the performance of network structure, and adopted a progressive growth method to gradually increase the number of blocks. However, due to the limitations of the LSTM model, the predictor they designed can only predict the performance of child models without skip-connections, which greatly limits its versatility.

Based on the above analysis, we believe that the above methods are only upgraded or improved in a certain aspect. When applied to practical work, certain methods will always make unnecessary compromises in search due to one or more defects. This is because the previous work only implemented the fingers or eyes of the Transformers, and we wanted to go one step further and turn the fingers or eyes into arms and heads. In summary, our contributions are as follows:

\begin{itemize}
\item We built an alternate neural network search framework to decouple the search space of skip-connection and operator. We abandoned the search framework of RL and adopted BOHB as the searcher to improve efficiency.
\item We designed a GCN-based(Graph Convolutional Network) predictor to predict performance with a Skip-Connection child model.
\item We design a generalized search space, which not only contains the traditional 13 candidate operators, but also adds the parameter initialization method and activation function of operator-level, and combines them in the form of Cartesian product.This expanded the search space, but still kept the search performance at a high level thanks to our "dual accelerators" and decoupling Settings.
\end{itemize}

\section{Related Work}
Recent years, more and more studies have focused on neural network architecture search. NASnet \cite{zoph_neural_2016,zoph2018learning} applied reinforcement learning to NAS for the first time, and each searched network would be retrained and evaluated. ENAS\cite{pham2018efficient} solved the problem of retraining by using parameter sharing \cite{wang_apq_2020} for a class of pruning method, try to reduce the search space of redundancy \cite{liu2018darts,chen_progressive_2019} attempts to weaken the discrete space for continuous space, and adopts the method based on gradient optimization search \cite{falkner_bohb_2018,zela_towards_2018} built a searcher based on bayesian method, and uses a hyperband method to optimize the allocation of resources. Most of the above studies are based on some method to improve, and they are often limited in practical application.

Our work draws on and improves upon previous work. Zoph\cite{zoph2018learning} proposed a basic operator search space on which much of the current work has been implemented. Jiang\cite{jiang_neural_2019} built an alternate search framework based on RL, decoupled the search space of skip-connection and operator, and added regularization on this basis to reduce network overfitting. Deng\cite{deng_peephole_2017} used an LSTM-based predictor to predict the performance of the sampled subnetworks, and built a coding system of operators for operator embedding. Falkner\cite{falkner_bohb_2018} combined Bayesian optimization with Hyperband method to realize a highly robust neural network architecture searcher, BOHB. Our method combines the advantages of the above work and improves the disadvantages: alternating search and search space decoupled root can avoid the low search efficiency caused by different search paths. In order to further improve the efficiency, we replace the dual RL architecture proposed by Jiang\cite{jiang_neural_2019} with two BOHBs. Predictor can catalyze better network architecture, and we have designed a new Predictor that addresses the problem of not being able to predict the performance of models with complex topologies. Finally, we expanded the search space by adding activation functions and initialization methods.

\section{Methods}
\subsection{Search Space}
Based on the work of Jiang\cite{jiang_neural_2019}, we decouple the search space of skip-connection and operator.

Chen\cite{chen_progressive_2019} proved that NAS is more inclined to build a model with multiple skip-connections, because skip connections can make the gradient drop faster, which is considered to be a kind of overfitting of the searcher. Therefore, when dealing with the search space of skip-connection, we adopted the same method as Chen\cite{chen_progressive_2019}, adding a dropout for each skip-connection as a blocker, so that the searcher can explore more connection modes. Also, the Dropout rate will decrease as the number of searches increases.

We assume that both the activation function and the initialization method interact with the operator, so we couple the operator with the activation function and the initialization method into a generalized operator, whose search space is the generalized search space.Specifically, we have added the following initialization methods and activation functions:

\textbf{Initialization Methods}: $Uniform, Gauss, Gauss^{2}$

\textbf{Activation Functions}: $Selu, Relu, Elu, Tanh$

Each operator is combined with an activation function and an initialization method, respectively, to form a generalized operator. That is, the new search space is the Cartesian product of the three domains, which is 13 * 3 * 4.
\subsection{Bayesian Optimization and Hyperband}
In the selection of the searcher, we use Bayesian Optimization and Hyperband (BOHB), which combines the advantages of BO and Hyperband algorithms. Like HyperBand, BOHB uses a simple and coarse-grained observation to acquire knowledge of network structure and hyper-parameters, and relies on this knowledge to evaluate different budgets and allocate more resources to good configurations. The traditional Hyperband method uses uniform random sampling, which leads to inefficiency. The BOHB can use the Bayesian method to model the interaction between parameters using multi-core density estimation, and obtain the samples with the highest expected improvement.
As we know, Bayesian optimization is to sample network parameters by means of maximizing the acquisition function:
\begin{equation}\label{eq:ei}
  a(x) = \int \max(0, \alpha - f(x))dp(f|D).
\end{equation}
Where $D = \{(x_0, y_0), \dots, (x_{i-1}, y_{i-1})\}$ is the observable pool, namely the child model hyper-parameters and the accuracy of verification set, $\alpha$ is the acquisition function, $f$ is the objective function of $x_{i}$, and $p$ is the density estimation of $f$ under the condition of $D$.
We use TPE(Tree Parzen Estimator) to model the distribution of the objective function $f$:
\begin{equation}
	\begin{aligned}
		l(x) &= p(y < \alpha | x, D) \\ 
		g(x) &= p(y > \alpha | x, D)
	\end{aligned}
	\label{eq:tpe_densities}
\end{equation}
To select a new candidate $x_{new}$ to evaluate, it maximizes the ratio $\nicefrac{l(x)}{g(x)}$; Instead of using $p(f|D)$ to directly estimate $f$, we used TPE, because it can control the scale of model construction in linearity, which makes the modeling process stable, and the kernel estimation method has a good performance in the mixed space.
In order to establish the model more quickly, we adopted a multi-core density segmentation method:
\begin{equation}
\begin{aligned}
	N_{b, l} &= \max(N_{min}, q\cdot N_b)\\
	N_{b, g} &= \max(N_{min}, N_b - N_{b, l})
\end{aligned}
\label{eq:kde_split}
\end{equation}

Where $N_{min}$ is the smallest model hyper-parameter, $q$ is a discount rate. We use the BOHB algorithm with TPE to implement an alternate search engine for searching the search space $D_{sk}$and $D_{op}$ of skip-connection and operator respectively, and its pseudocode is shown in Algorithm \ref{alg:1}
\begin{algorithm2e}[htb]
\caption{sampling in Alternate BOHB}\label{alg:1}
        \DontPrintSemicolon
        \SetKwInOut{Input}{input}\SetKwInOut{Output}{output}
	\Input{search space $D_{sk}$ and $D_{op}$, drop rate $\rho$, discount rate $q$, number of samples $N_s$, minimum number of hyperparamaters $N_{min}$ to build a model, and bandwidth factor $b_w$}
        \Output{A combination of skip-connections and operators}
    \For{$D \in \{ D_{sk}, D_{op}\}$}{
        \lIf{rand() \textless $\rho$}{\Return random $x_{D}$}

	$b = argmax \left\{ D_b:  \vert D_b \vert \geq N_{min} + 2\right\}$\; 
        \lIf{ $b=\emptyset$}{ \Return random $x_{D}$}

	fit KDEs according to Eqs.~\eqref{eq:tpe_densities} and ~\eqref{eq:kde_split}\;
	draw $N_s$ samples according to $l'(x)$ (see text)\;
        sample $x_{D}$ with highest ratio  $l(x)/g(x)$
        }
    \Return \{$x_{D_{sk}}, x_{D_{op}}$\}

\end{algorithm2e}
\subsection{Predictor}
\begin{figure}
    \centerline{\includegraphics{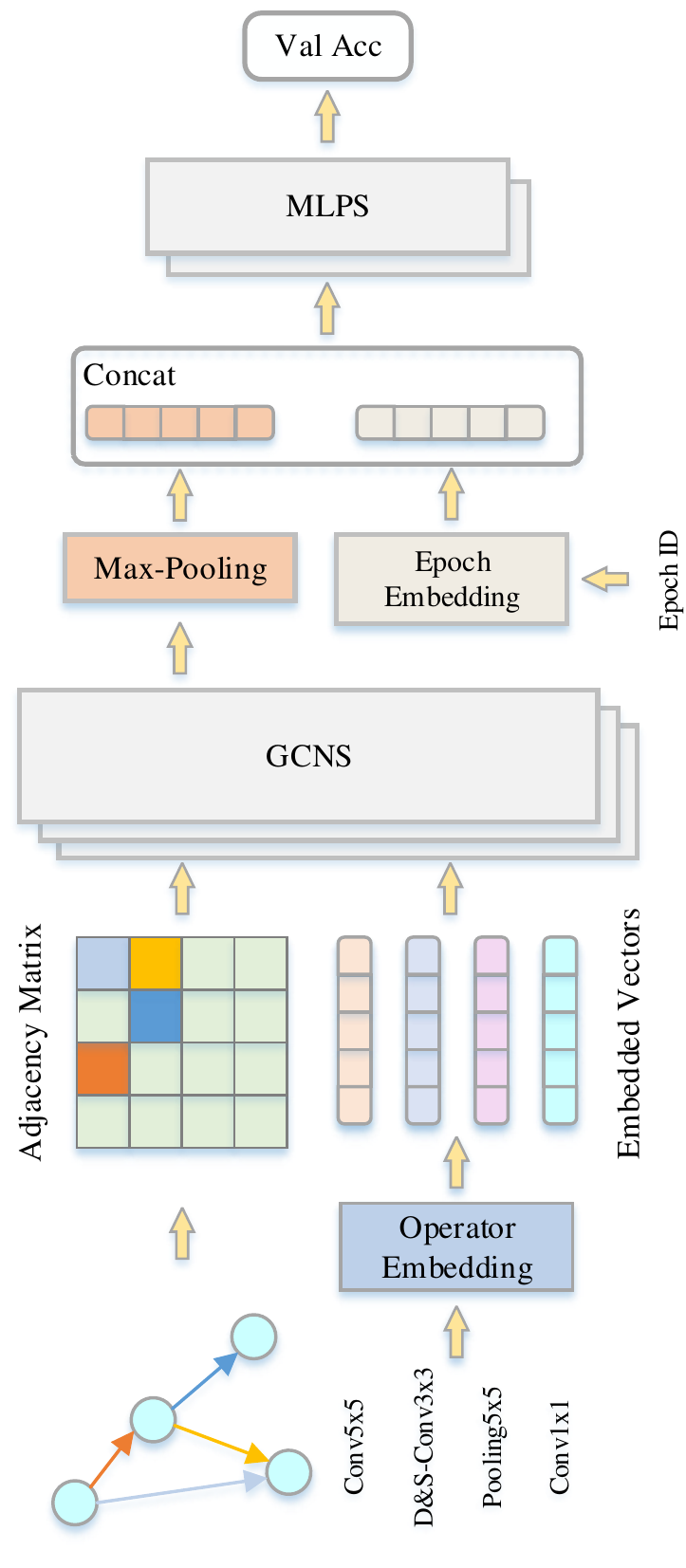}}
	\caption{Graphical Predictor Architecture}
	\label{fig:predictor}
\end{figure}
\textbf{Model Structure:} Similar to Deng\cite{deng_peephole_2017}  and Liu\cite{liu2018progressive}, we designed a predictor to predict the performance of the child model. But in the previous study,  was unable to predict the performance of a network structure with a complex topology, such as applying skip-connection to a network. Zoph\cite{zoph_neural_2016} mentioned a skip-connection representation method inside the controller. Their model abstracted skip-connection into DAG, and added $n-1$ "anchor points" wrapped by Sigmoid at the end of the representation of the $n$-th network node to sample whether the layer is connected with the previous layer. Abtracting skip-connection to a DAG results in more choices for the later layer and fewer for the front layer, which we believe is unfair in predictor. Therefore, in our predictor, connections are assumed to be bi-directional (such connections form an undirected graph). The network structure can then be represented by a symmetric adjacency matrix, where each point in the graph contains an operator that is encoded to characterize the node. Based on the above analysis, we construct a GCN network as graphical predictor to predict the accuracy of the child model on the verification set:
\begin{align}
    H^{(k)}&=M\left(\mathbf{A}, \mathbf{H}^{(k-1)} ; \mathbf{W}^{(k)}\right)\\ \label{eq:1}
    &=\operatorname{Relu}\left(\tilde{\mathbf{D}}^{-\frac{1}{2}} \tilde{\mathbf{A}}\tilde{\mathbf{D}}^{-\frac{1}{2}} \mathbf{H}^{(k-1)}\mathbf{W}^{(k-1)}\right)\\\label{eq:2}
    output&=\operatorname{MLPs}\left (\left [\max \left (\mathbf{H}^{(l)}\right ); \mathbf{E} \right ]\right ) 
\end{align}
Where $\mathbf{H}^{(0)}=\mathbf{X}$, $\mathbf{A}$ is the network structure of adjacency matrix, Obtained by sampling the search space of skip-connection, $\tilde{\mathbf{A}} = \mathbf{A} + \mathbf{I}$, $\mathbf{I}$ is the identity matrix, $\mathbf{W}^{(\ast)}\in\mathbb{R}^{m\times n}$ is a trainable parameter, $\tilde{\mathbf{D}}^{(ii)} = \sum_{j}\tilde{\mathbf{A}}^{(ij)}$ is the degree matrix of matrix $\mathbf{A}$, $\mathbf{X}$ and $\mathbf{E}$ are respectively the embedding of operators and epoch:
\begin{align}
    \mathbf{X} =\operatorname{Embedding}\left(\mathbf{e}_{op}\right );     \mathbf{E} =\operatorname{Embedding}\left(\mathbf{e}_{ep}\right )
\end{align}
GCN of (\ref{eq:1}) is used to extract features of network structures and operators, and is finally represented by $\mathbf{H}^{(l)}$ after max-pooling, which is called topological feature. We concatenate the topology features and the number of iterations after embedding, and then make the accuracy prediction through a 3-layer MLPs with $BN$ and $ReLU$.

\textbf{Training Objective:} By alternating sampling with BOHB, we can obtain a set of child models${\{(s_{i}, o_{i})\}}_{1:n}$. After these networks are trained, predictor receives a coded set of network profiling parameters and a verification set of accuracy${\{(A_{i}, X_{i}, E_{i}), \alpha_{i}\}}_{1:n}$. Therefore, predictor can perform supervised learning just like any other machine learning model. Specifically, we use MSE as the loss function:
\begin{equation}
    \mathcal{L}(\mathcal{D} ; \boldsymbol{\theta})=\frac{1}{N} \sum_{i=1}^{n} L\left(f\left(A, X, E\right)_{i}, y_{i})\right)
\end{equation}
Where, $\theta$ represents the parameters of the predictor. Note that since we use the BOHB sampling network, the number of iterations allocated to different networks is different, but we all use the accuracy rate obtained from the last iteration as the label.

By sampling a lot, graphical predictor can also, like the HB algorithm, identify the effects of different resources on different network structures, which may help the network find a more appropriate number of iterations, a kind of early stop. Of course, the more important role of predictor is to determine the performance of different network structures, and with that in mind, you can also find what predictor considers to be better structures by factoring the adjacency matrix.

\subsection{GPNAS}
\begin{figure}[htb]
	\centerline{\includegraphics[width=0.5\textwidth]{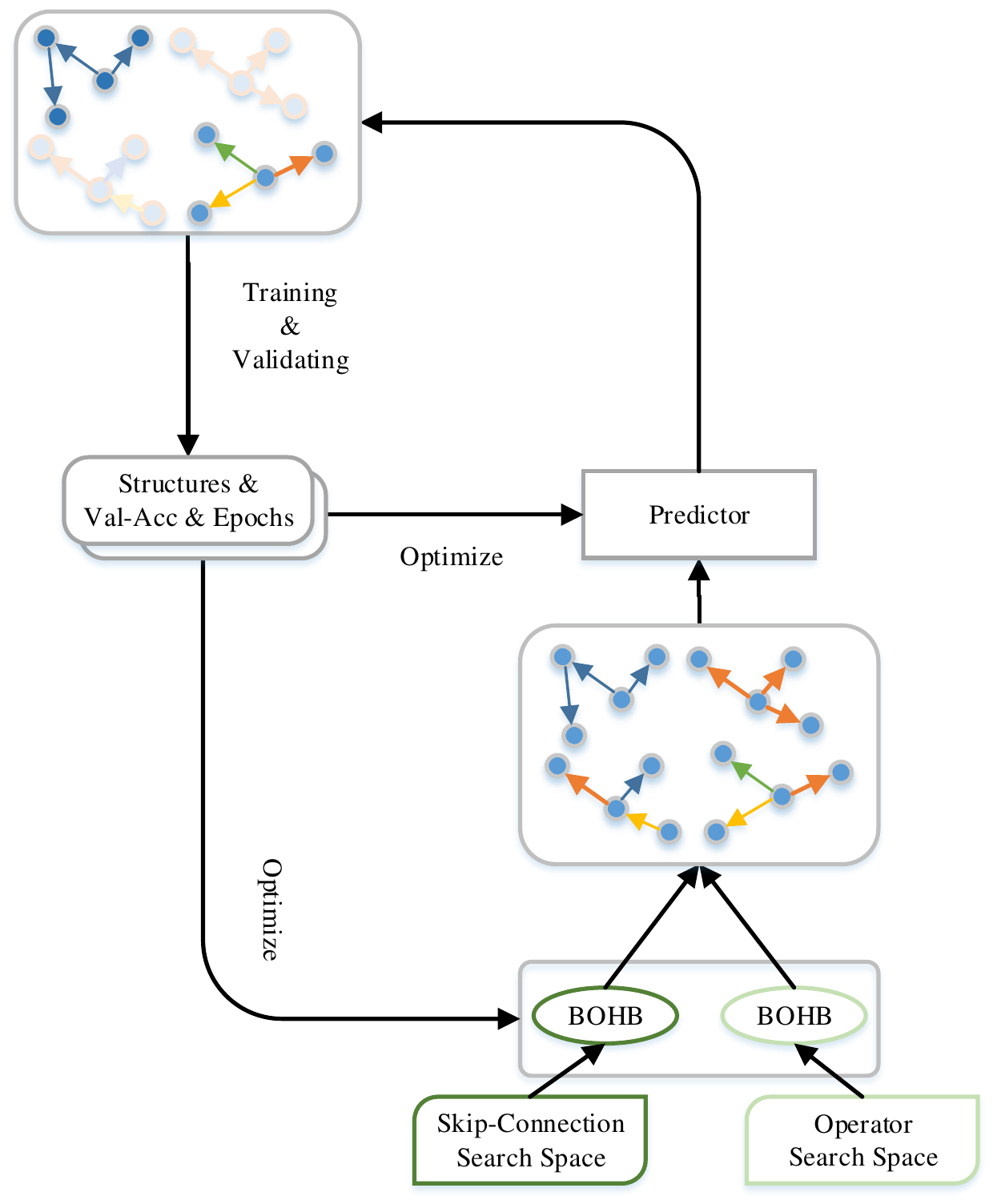}}
	\caption{GPNAS framework workflow. BOHB samples skip-connections and operators from the two search Spaces to form a child model, and then predicts performance using predictor. If predictor decides that the child model is worth training, it trains and evaluates the child model, and then feeds back to predictor and the BOHB to optimize both.}
	\label{fig:GPNAS}
\end{figure}
Through above components, we build a NAS framework based on the generalized search space. The BOHB samples the search space of skip-connection and predictor, respectively, and combines the two parameters to form a child model. In the early stage of search, BOHB samples a large number of sub-models in parallel and trains them. The structure of the child models, the hyper-parameters, the accuracy of the validation set, and the number of iterations will make up predictor's training set. At a later stage, predictor will act as a filter to help the BOHB find high-quality networks faster, creating a dual accelerator with the BOHB. The specific process is shown in Figure \ref{fig:GPNAS}.

\section{Experiments}
\subsection{Datasets}

\textbf{Image Classification on CIFAR-10}
CIFAR-10 is a color image data set closer to universal objects. CIFAR-10 is a small data set compiled by Hinton students Alex Krizhevsky et al. for identifying universal objects. A total of 10 categories of RGB color pictures: airplane , automobile , bird, cat, deer, dog, frog, horse, ship and truck.
The size of each picture is 32×32, each category has 6000 images, and there are 50,000 training pictures and 10,000 test pictures in the dataset.

\textbf{NAS-Bench-101\cite{Ying2019NAS-Bench-101}}
In order to really evaluate a search algorithm and bypass the computational challenge, Christ et al. collected NAS-Bench-101, which enumerated all possible DAGs with nodes $\le$ 7 to form a (420k+) network and its final test accuracy.

\textbf{NAS-Bench-201\cite{NAS-Bench-201DongYang}}
NAS-Bench-201 contains all possible architectures generated by 4 nodes and 5 related operators, resulting in a total of 15,625 neural cell candidates.

\textbf{NAS-Bench-1Shot1\cite{Zela2020aNAS-Bench-1Shot1}}, a set of 3 new benchmarks for one-time neural architecture searches, allows us to cheaply track the trajectory and performance of found architectures.

\subsection{Relationship to PriorWork}

\textbf{DARTS}\cite{liu2018darts} weakens the search space into a continuous spatial structure, so that gradient descent can be used for performance optimization. DARTS can search for a high-performance framework with a complex graph topology in a larger search space.

The continuity of search space is not the original method, but Darts is different from the traditional method in that the traditional method is to search for a special structure (Filters Shape, Branch Pattern) in a continuous search space, while DARTS is to search for a complete network structure.

DARTS is excellent. It proposes a new paradigm based on gradient optimization. However, there are several problems:

$-$ Each output needs to use the weights of all connection layers and activation functions, and in order to be able to back-propagate, it is necessary to save a lot of intermediate results, so only a relatively small space can be searched;

$-$ Simple weighting cannot measure the contribution of different connections and activation functions, that is, the optimal solution may be missed in the search stage;

$-$ Transfer learning on different data sets is problematic. The search structure on small data sets requires manual splicing of large networks. Of course, this problem is not unique to DARTS;

$-$ The difference between the search stage and the inference stage is that all calculated weights are used when searching, and only the highest weight is retained when inferring;

$-$  Only search for cells, and then stack them up continuously. In fact, as long as the network is stacked deep enough, the performance will not be too bad;

$-$  The structural parameters of some operations are very close. For example, the structural parameters of two operations are 20.2\% and 20.1\% respectively. It seems that there is not much difference between choosing 20.2\% and 20.1\%;

$-$ In the beginning, training was performed on the entire Net, but when selecting only a sub-network, some problems may occur in the middle.

The progressive idea put forward by \textbf{PNAS}\cite{liu2018progressive} provides an important foundation for a large number of subsequent research work. PNAS designs a surrogate model to  search for the blocks inside the cells, and then evaluates the accuracy of the cells made up of new blocks on specific tasks in order to find the most suitable cells. Our method is the same as the original intention of PNAS, which is to search the blocks inside the cells to find the most suitable cells.

\textbf{SNAS}\cite{xie2018snas} is a neural network structure search (NAS) framework with high efficiency, low deviation and high degree of automation. By remodeling the NAS, the author theoretically bypassed the problem of slow convergence in the complete delay reward based on reinforcement learning methods, and directly optimized the objective function of the NAS through the gradient to ensure that the network parameters of the resulting network can be used directly.
Compared with other NAS methods that generate sub-networks based on certain rules, the author proposes a more automatic network topology evolution method, which optimizes the accuracy of the network while limiting the complexity and forward delay of the network structure.

\textbf{ProxylessNAS}\cite{cai2018proxylessnas} does not need to use proxy tasks, and directly searches the entire network on large-scale data and provides a new way of path pruning for NAS, showing the close relationship between NAS and model compression, and finally reducing memory consumption by an order of magnitude through binarization. ProxylessNAS proposes a gradient-based method (delay regularization loss) to constrain hardware indicators.

\subsection{Architecture Search Evaluation}
\begin{table*}[htb]
\caption{The comparisons of our method search results to
other state-of-the-art results on CIFAR-10.}
\begin{center}
\begin{tabular}{|l|c|c|c|c|}
\hline \textbf{Method} & \textbf{GPUs} & \textbf{Times (days)} & \textbf{Params (M)} & \textbf{Error (\%)} \\
\hline DenseNet-BC \cite{huang2017densely} & $-$ & $-$ & 25.6 & 3.46 \\
DenseNet + Shake-Shake \cite{Gastaldi} & $-$ & $-$ & 26.2 & 2.86 \\
DenseNet + CO \cite{DeVries} & $-$ & $-$ & 26.2 & 2.56 \\
\hline Budgeted Super Nets \cite{Veniat} & $-$ & $-$ & $-$ & 9.21 \\
ConvFabrics \cite{Saxena} & $-$ & $-$ & 21.2 & 7.43 \\
Macro NAS + Q-Learning \cite{Baker} & 10 & 8-10 & 11.2 & 6.92 \\
Net Transformation \cite{cai2018path} & 5 & 2 & 19.7 & 5.70 \\
FractalNet \cite{Larsson} & $-$ & $-$ & 38.6 & 4.60 \\
SMASH \cite{Brock} & 1 & 1.5 & 16.0 & 4.03 \\
NAS \cite{zoph_neural_2016} & 800 & 21-28 & 7.1 & 4.47 \\
NAS + more filters \cite{zoph_neural_2016} & 800 & 21-28 & 37.4 & 3.65 \\
AmoebaNet-A + Cutout \cite{real2018regularized} & 3150 & 1 & 3.2 & 3.34 \\
AmoebaNet-B + Cutout \cite{real2018regularized} &3150& 1& 2.8 & 2.55\\
PNAS \cite{liu2018progressive}&225&1&3.2&3.41\\
\hline Hierarchical NAS \cite{liu2017hierarchical} & 200 & 1.5 & 61.3 & 3.63 \\
Micro NAS + Q-Learning \cite{zhong2018practical} & 32 & 3 & $-$ & 3.60 \\
Progressive NAS \cite{liu2018progressive} & 100 & 1.5 & 3.2 & 3.63 \\
NASNet-A \cite{zoph2018learning} & 450 & 3-4 & 3.3 & 3.41 \\
NASNet-A + Cutout \cite{zoph2018learning} & 450 & 3-4 & 3.3 & 2.65 \\
DARTS  + Cutout \cite{liu2018darts} & 4 & 1  & 3.3   & 2.76 \\ 
SNAS + Mild Constraint + Cutout\cite{xie2018snas}  & 2 &  1 &  2.9  & 2.98\\ 
SNAS + Moderate Constraint + Cutout\cite{xie2018snas}  & 2 & 1  &  2.8  & 2.85\\ 
SNAS + Aggressive Constraint + Cutout\cite{xie2018snas}   & 2 & 1  & 2.3   & 3.10\\ 
  P-DARTS CIFAR10 + Cutout \cite{chen_progressive_2019} & 0.3 &  1 &  3.4  & 2.50 \\ 
ENAS  + Cutout \cite{pham2018efficient} & 1 & 0.45 & 4.6 & 2.89\\
\hline 
GPNAS& 280 & 1 & 3.3 & 2.78\\
\hline
\end{tabular}
\label{tab:results}
\end{center}
\end{table*}

We use 280 NVIDIA 1080Ti to perform neural network architecture search on CIFAR-10. We use one GPU as the server, and the rest as the client. In order to further speed up the evaluation of the neural network, we terminated the training early in the 50th time period of the pre-training period. And selected the top 20 networks from the pre-training, and fine-tuned them for 300 periods to obtain the final accuracy.

For ImageNet training, we use the same RCell and NCell defined on ZIFAR10 in accordance with the accepted standard (ie mobile settings) to build the network\cite{zoph_neural_2016}.

Table \ref{tab:results} summarizes the SOTA results on CIFAR10, and GPNAS achieves SOTA accuracy with fewer samples. Due to early termination and transfer learning, our end-to-end search cost (that is, GPU days) is also comparable to the SOTA method. What is pleasantly surprised is that the accuracy of GPNAS and DARTS are similar. For the case of using cutout, the sample is reduced by 6 times.

\subsection{Sample Efficiency}
In order to further check the sample efficiency of the model, we conduct an evaluation on GPNAS on the latest NAS benchmark evaluation data set NAS-Bench-101\cite{Ying2019NAS-Bench-101}. NAS-Bench-101 enumerates all possible DAGs with nodes $\le$ 7, these DAGs are composed of (420k +) networks, and finally verify their accuracy. In this way, computing resources can be greatly reduced, and sample efficiency can be evaluated fairly.


On NAS-Bench-101, we conducted 200 trials on three algorithms (GPNAS, RE and RS). The search goal is to find the neural network with the highest average test accuracy (globally optimal). We choose Random Search (RS)\cite{Sciuto2019} and Regularized Evolution (RE) \cite{real2018regularized} as the evaluation benchmark algorithms, because RE provides competitive results in the following experiments.

\begin{figure*}[htb]
  \centering\small
  \begin{minipage}[t]{0.5\linewidth}
    \centering
    \includegraphics[width=\textwidth]{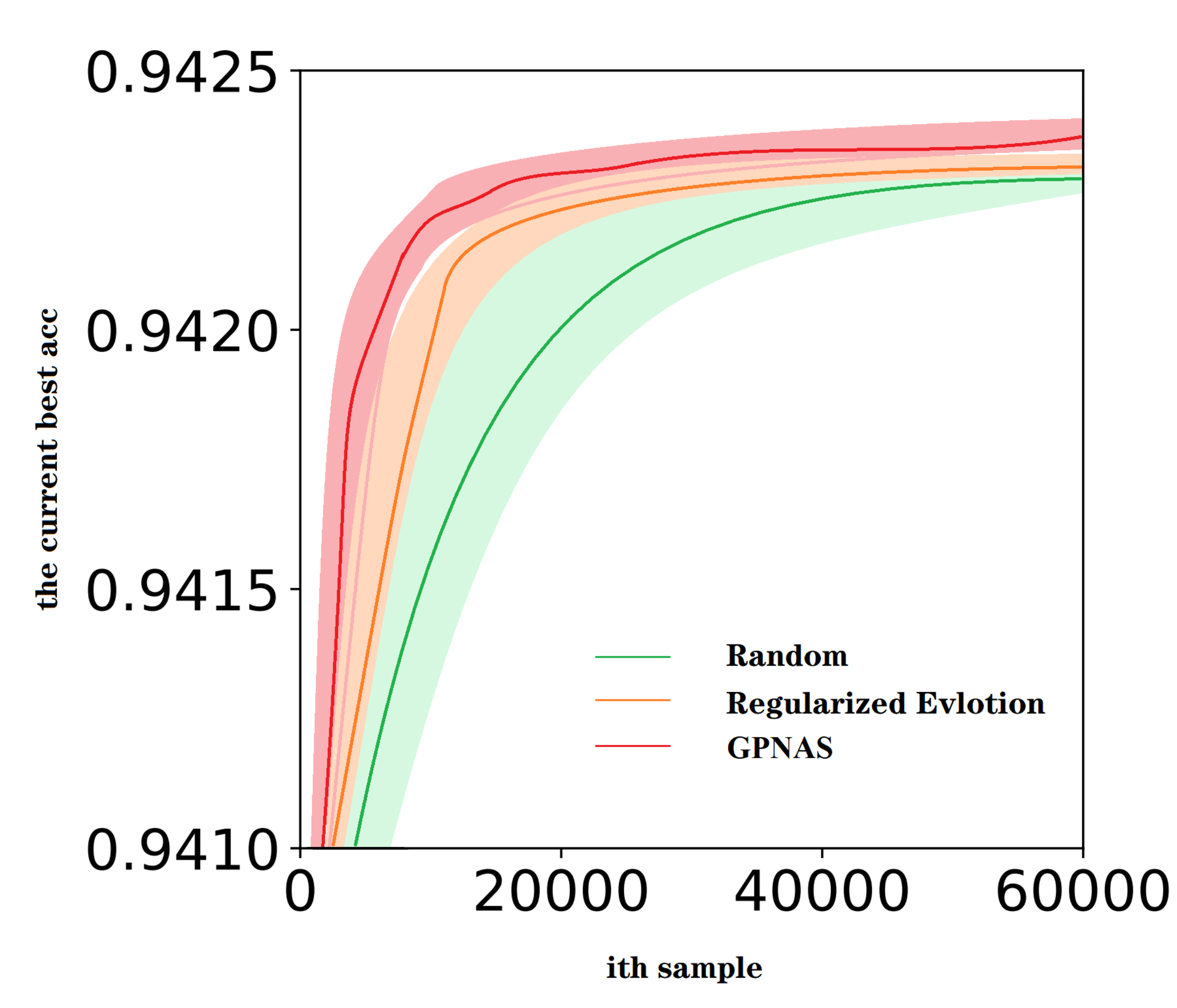}
   (a) best acc progression
  \end{minipage}%
  \begin{minipage}[t]{0.5\textwidth}
    \centering
    \includegraphics[width=\textwidth]{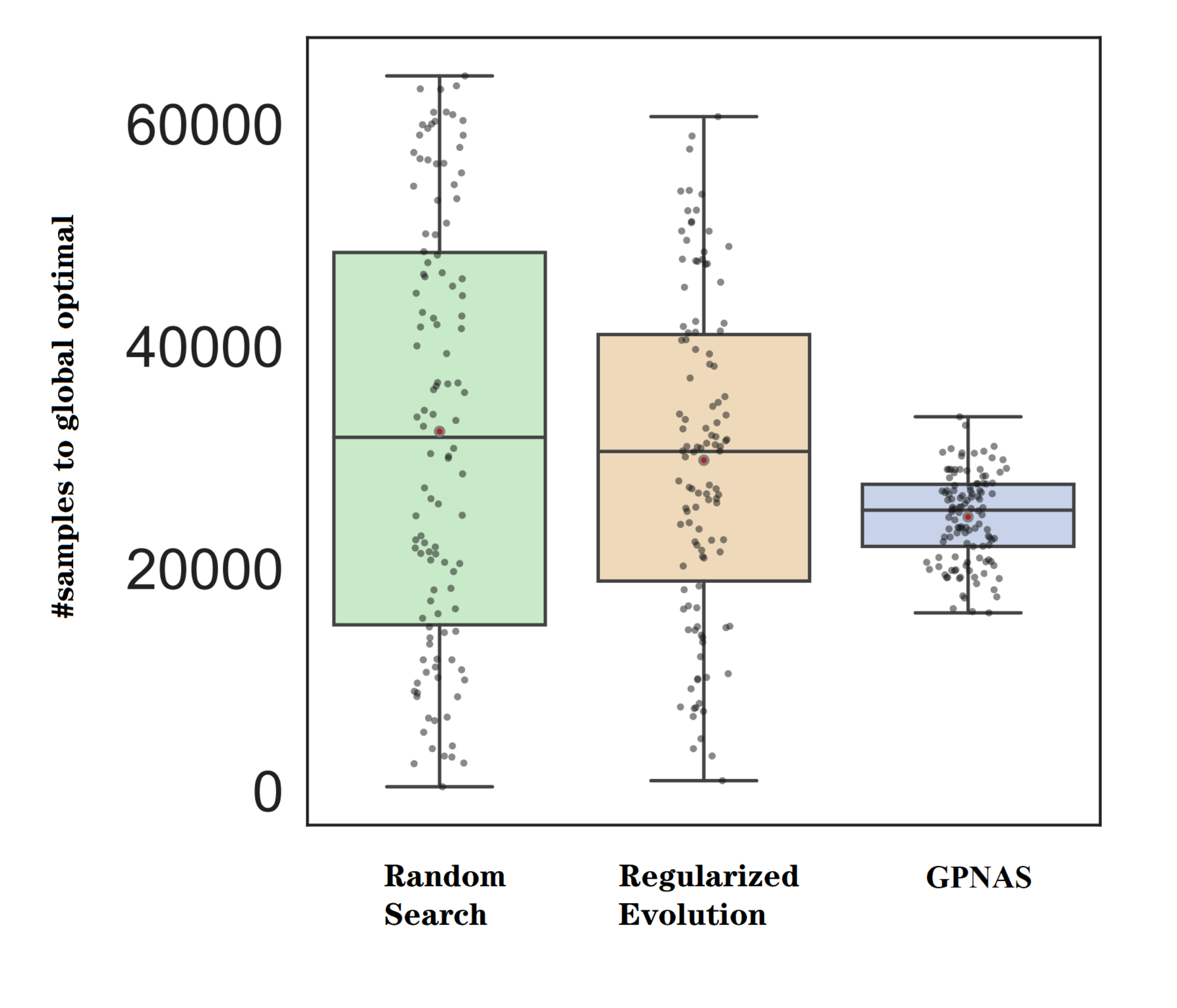}
    (b) \#samples to the best
  \end{minipage}  \\
  \caption{Finding the global optimum on NASBench-101:
GPNAS is 5x, 3x faster than Random Search and Regularized Evolution on NAS-Bench-101 (nodes $\le$ 6). The results
are from 200 trails with different random seeds.} 
\label{fig:performance}
\end{figure*}
Figure \ref{fig:performance} demonstrates that GPNAS is 3 and 5 times faster than RE and RS, respectively. Regularized evolution only occurs genetic mutations on the top k models with the highest performance. It should be noted that the subtle differences in Figure 7a actually reflect the huge speed gap shown in Figure 7b. From Figure 1 we can also see that there are a large number of architectures in the generalized search space we constructed, and the difference between its performance and the global best performance is small. Therefore, finding the top 5\% of the architecture is very fast, but the process of reaching the global optimum is very slow.
\begin{figure*}[htb]
  \centering\small
  \begin{minipage}[t]{1\linewidth}
    \centering
    \includegraphics[width=\textwidth]{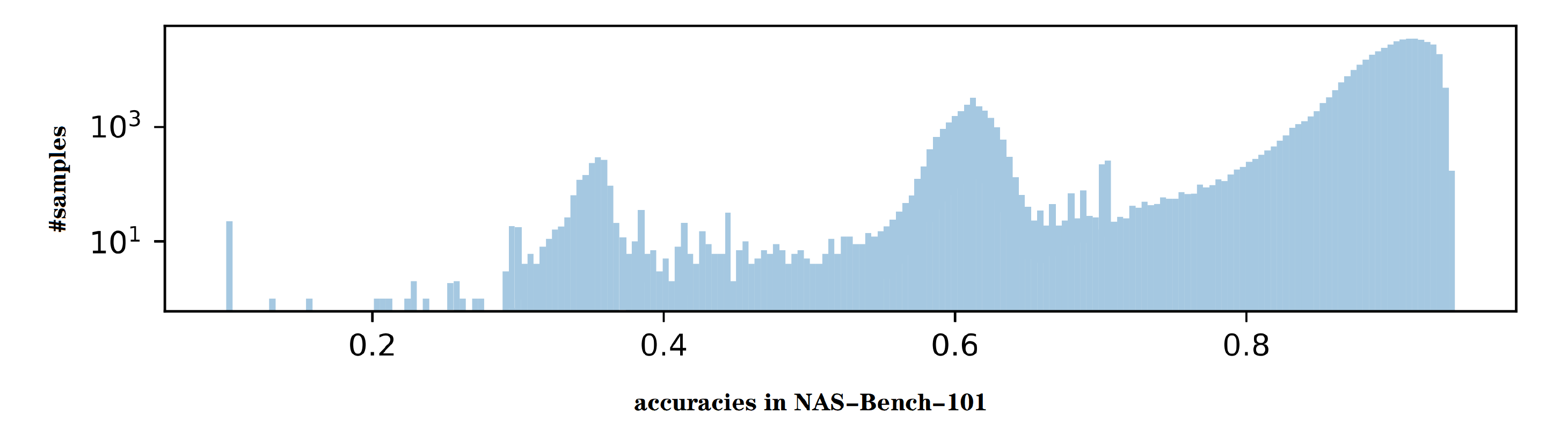}

  \end{minipage}%

  \caption{Accuracy distribution of NASBench.}
  \label{fig:acc}
\end{figure*}

Figure \ref{fig:acc} shows the change in accuracy of the graphical predictor on NAS-Bench-101 as the sample size increases.
\subsection{Cell Graph Structure Stability}
\begin{figure}[htb]
	\centerline{\includegraphics[width=0.5\textwidth]{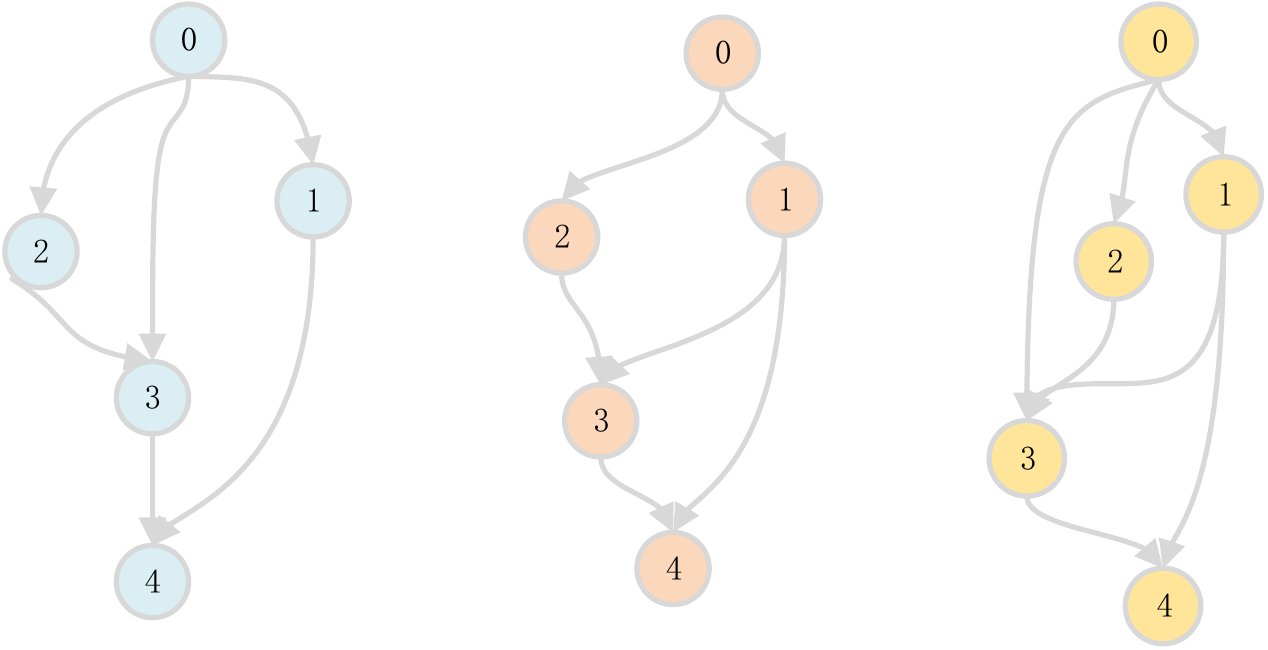}}
	\caption{Operator-stable cell graph structure found by GPNAS with 5 operators (left), 6 operators (middle) and 7 operators (right).}
	\label{fig:cell_graph}
\end{figure}
Theoretically, we can analyze the advantages of different network structures by using graphical predictor, but the model performance is expressed by the cell formed after the structure and operation coupling.  Due to the above limitations, we designed a set of experiments that fixed model structure, like NASnet. and only ran BOHB of operator. Operator BOHB run under 10 different cell graphs (the optimal structure mentioned in previous NAS studies, including GPNAS) and are sampled for 24 GPU days respectively, and graphical predictor is used to predict the average error of the model that's made up of these cells.
\begin{table*}[htb]
\caption{The stability of the optimal cell structure under the influence of different operators found in 10 different studies.}
\begin{center}
\begin{tabular}{|l|c|c|c|c|}
\hline \textbf{Method} & \textbf{Operators} & \textbf{Params (M)} & \textbf{Avg-Error (\%)} & \textbf{Top-Error (\%)}\\
\hline AmoebaNet-A \cite{real2018regularized} & 7-10 & 2-4 & 15.84 & 3.34\\
AmoebaNet-B \cite{real2018regularized} & 7-10 & 2-4 & 14.55 & 2.55\\
AdvantageNAS \cite{sato_advantagenas_2020} & 5-6 & 1-2 & 18.32 & 2.98\\
\hline SNAS + Moderate Constraint \cite{xie2018snas} & 4-8 & 2-4 & 13.62 & 2.85\\
SNAS + Aggressive Constraint \cite{xie2018snas} & 3-5 & 2-4 & 13.54 & 3.10\\
DARTS \cite{liu2018darts} & 4-8 & 2-4  & 14.25   & 2.76 \\ 
P-DARTS \cite{chen_progressive_2019} & 4-8 &  3-4 & 14.71  & 2.50 \\
\hline NAS \cite{zoph2018learning} & 6-8 & 21-28 & 16.91 & 4.47\\
ENAS \cite{pham2018efficient} & 6-10 & 0.2-1 & 19.33 & 2.89\\
\hline 
GPNAS& 5-7 & 2-4 & 13.88 & 2.78\\
\hline

\end{tabular}
\end{center}
\label{tab:stability}
\end{table*}
Table \ref{tab:stability} shows that a low error rate can still be achieved by sampling the operator after using the optimal structure found. However, fewer parameters (ENAS and AdvantageNAS) and more parameters (NAS) performed lower stability.This may be related to the depth of the network, where overfitting is more likely to occur at larger depths, and the effect of operators is magnified at smaller depths.

\subsection{Validation Regret and Test Regret }
Next We evaluated the performance of GPNAS on four NAS benchmarks and plotted the validation regret:NAS-Bench-101\cite{Ying2019NAS-Bench-101}, NAS-HPO\cite{KleinHutter}, NAS-Bench-1shot1\cite{Zela2020aNAS-Bench-1Shot1} and NAS-Bench-201\cite{NAS-Bench-201DongYang}. We compare against several baseline algorithms, namely Random Search (RS)\cite{bergstra_random_2012}, BOHB\cite{falkner_bohb_2018}, Tree Parzen Estimator (TPE) \cite{Bergstra}, Hyperband (HB)\cite{li_hyperband_2018} and regularized evolution (RE)\cite{real2018regularized},Differential Evolution \cite{Awad}.

We consider BOHB as the main baseline algorithm (running to 10Ms) because it belongs to the same algorithm family as GPNAS and has been proven to be reliably executed many times before. For each algorithm, we performed 500 independent runs and reported the average performance of verification regret Ying\cite{Ying2019NAS-Bench-101}. Throughout the entire process, we evaluate the algorithm anytime and anywhere, showing the performance of the best configuration over time, as suggested by Ying\cite{Ying2019NAS-Bench-101} and Lindauer\cite{Lindauer}. In all our graphs, the x-axis shows the estimated clock time, which is the cumulative time spent training each architecture found by the NAS benchmark.
\begin{figure*}[b]
  \centering\small
  \begin{minipage}[t]{0.5\linewidth}
    \centering
    \includegraphics[width=\textwidth]{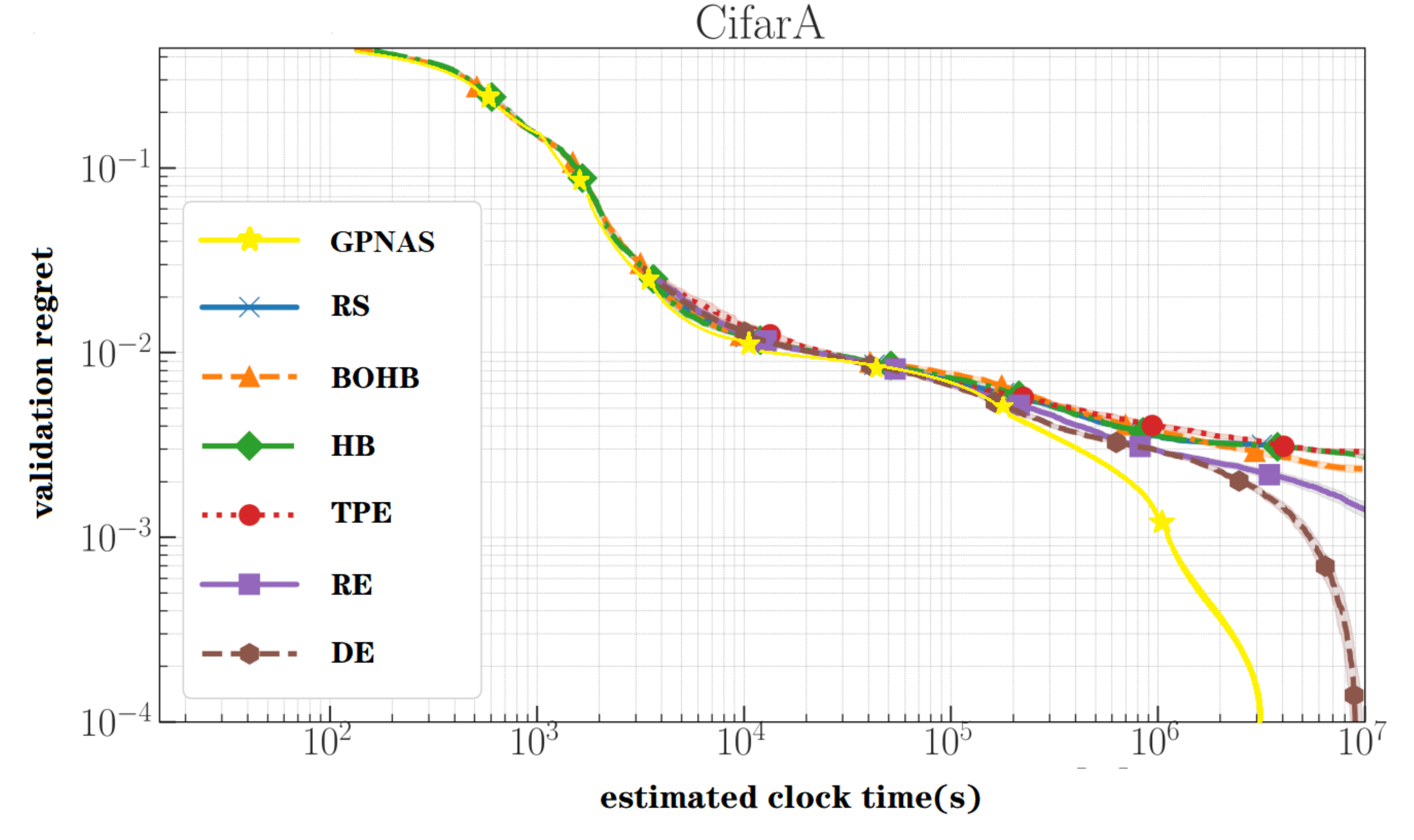}
   (a)  NAS-Bench-101 on CifarA
  \end{minipage}%
  \begin{minipage}[t]{0.5\textwidth}
    \centering
    \includegraphics[width=\textwidth]{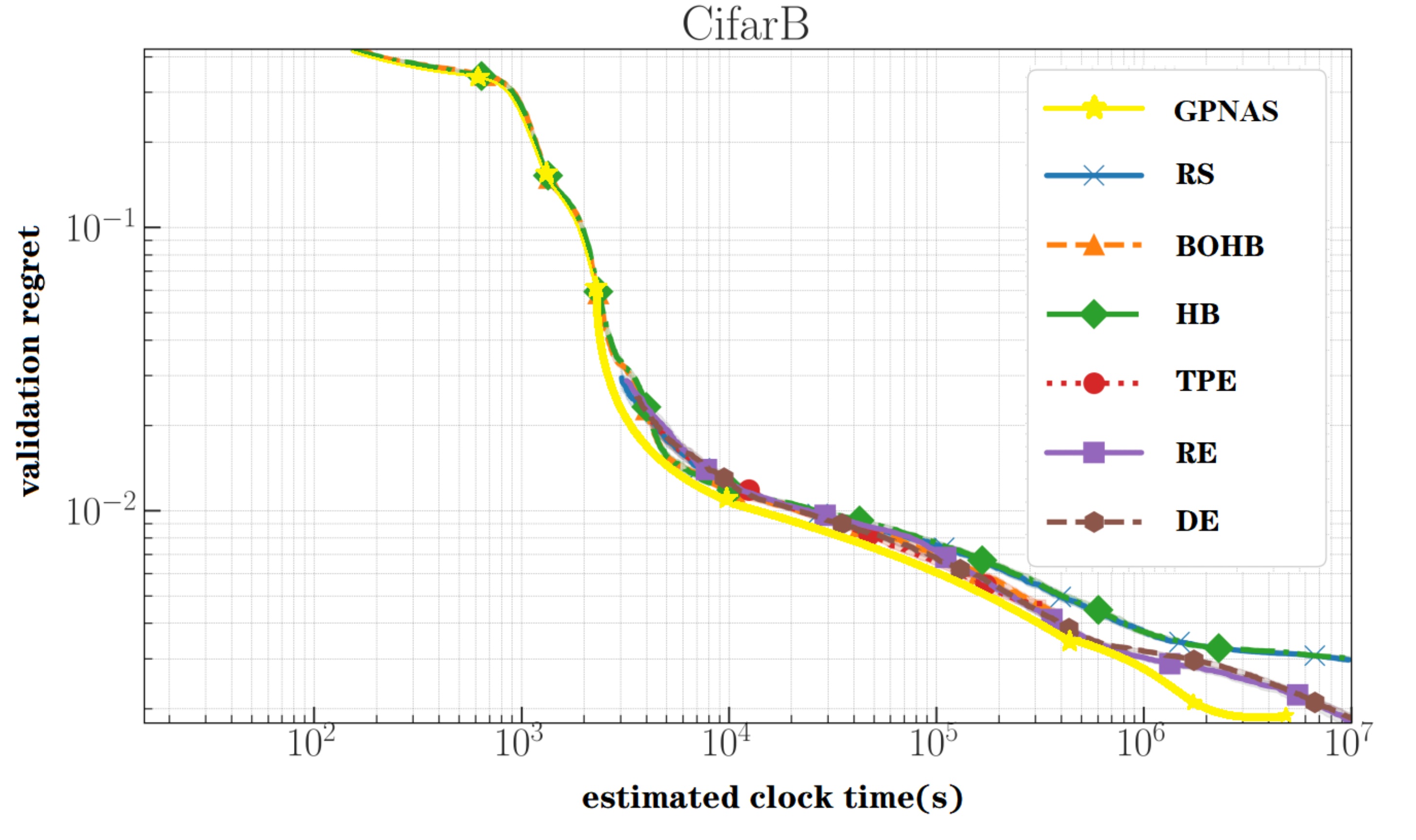}
    (b)  NAS-Bench-101 on  CifarB 
  \end{minipage}  \\
  \begin{minipage}[t]{0.5\textwidth}
    \centering
    \includegraphics[width=\textwidth]{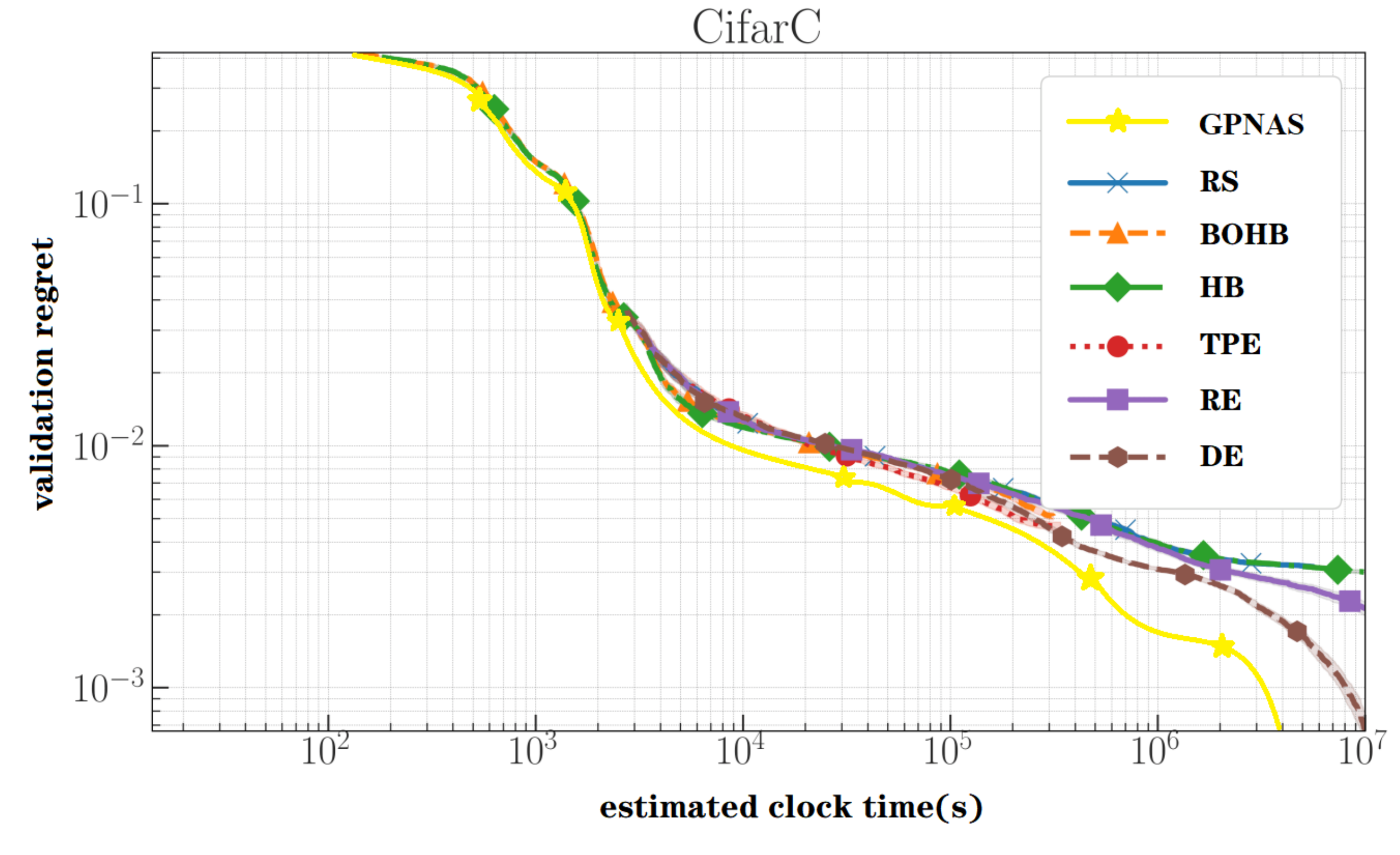}
    (c) NAS-Bench-101 on  CifarC
  \end{minipage}%
  \begin{minipage}[t]{0.5\textwidth}
    \centering
    \includegraphics[width=\textwidth]{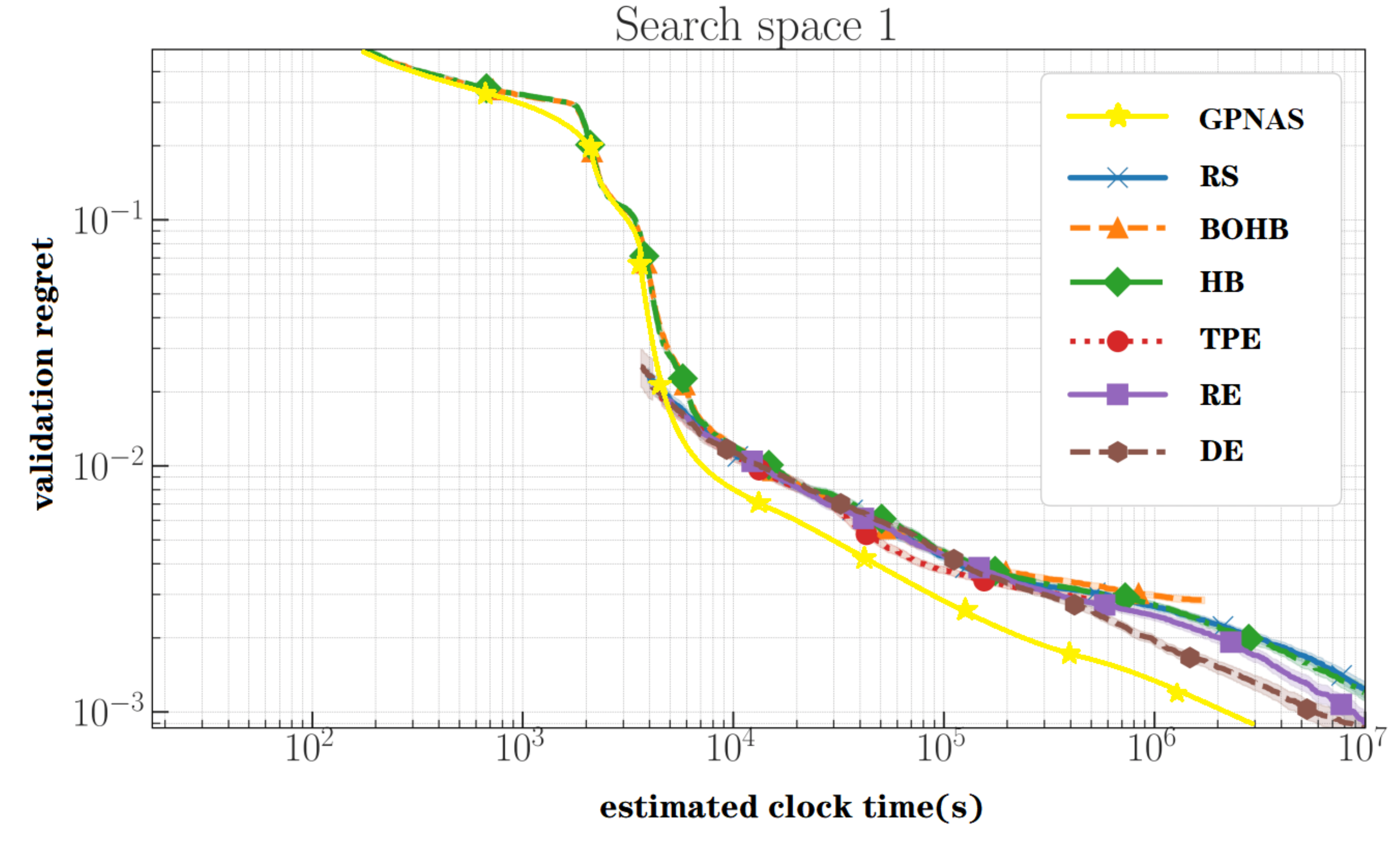}
    (d)  NAS-1Shot1 on the Search Space 1 
  \end{minipage}
    \begin{minipage}[t]{0.5\textwidth}
    \centering
    \includegraphics[width=\textwidth]{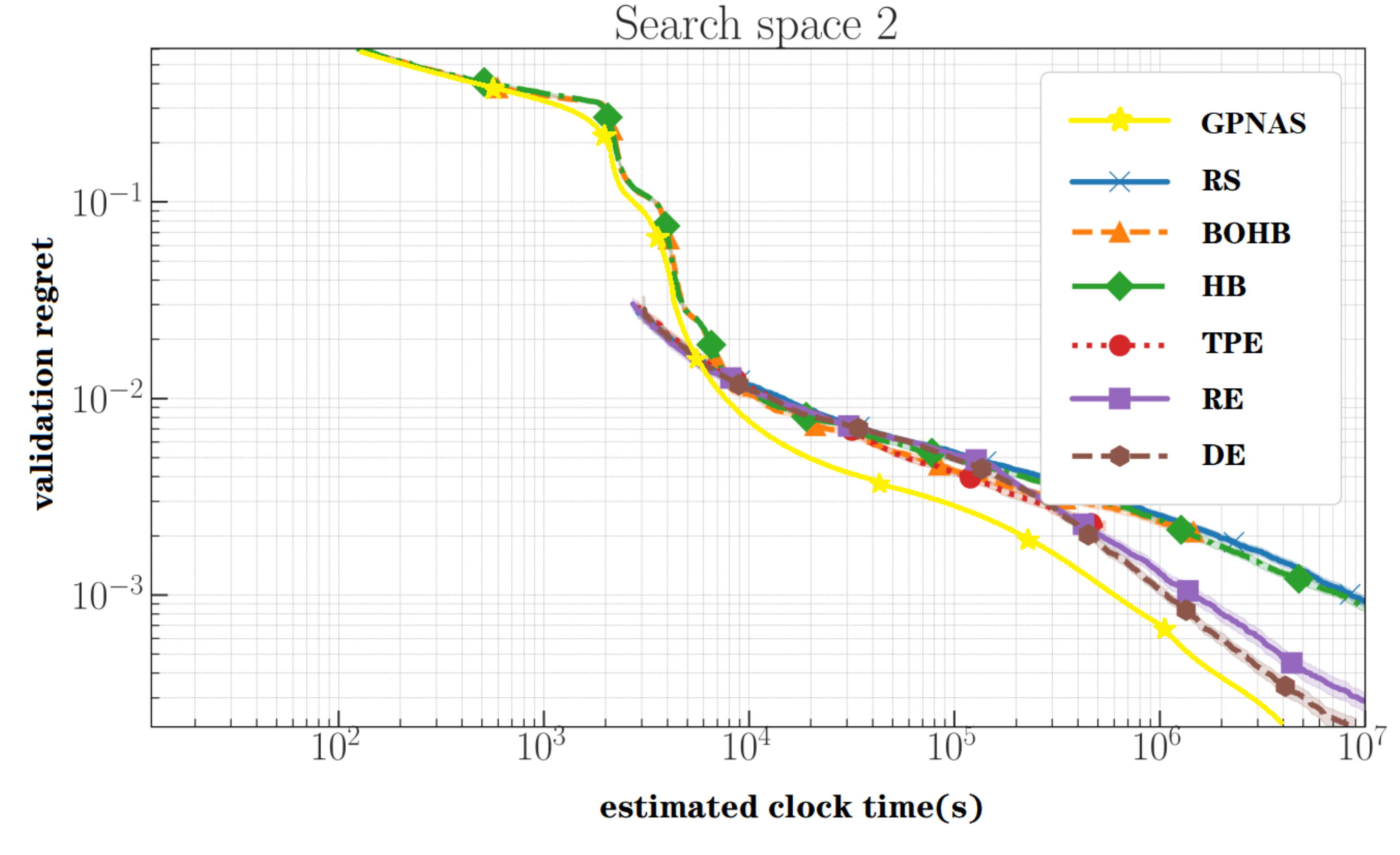}
    (e) NAS-1Shot1 on the Search Space 2 
  \end{minipage}%
  \begin{minipage}[t]{0.5\textwidth}
    \centering
    \includegraphics[width=\textwidth]{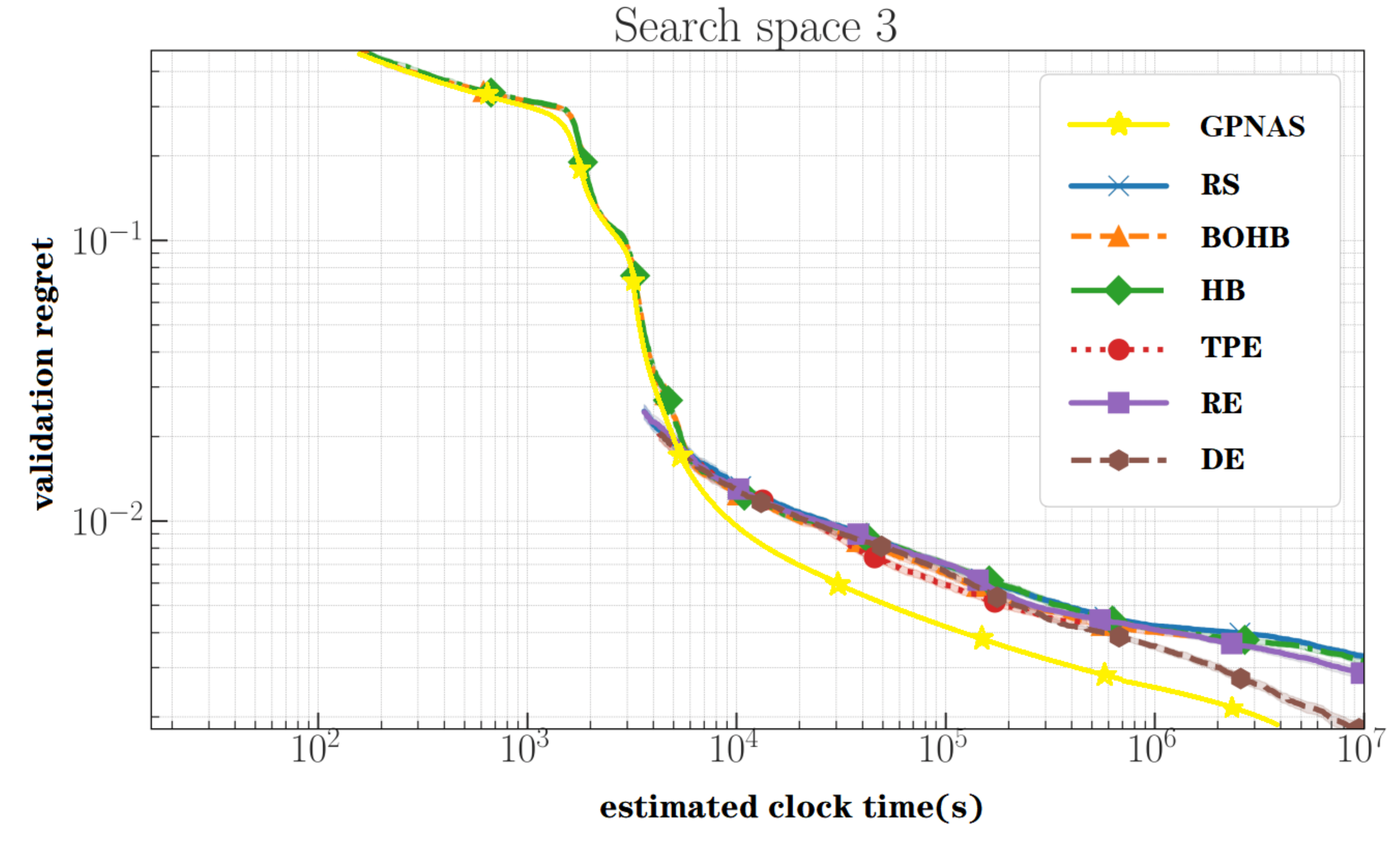}
    (f) NAS-1Shot1 on the  Search Space 3
  \end{minipage}
  \caption{A comparison of the mean validation regret performance of 500 independent runs as a
function of estimated training time for NAS-Bench-101 on the the three different CIFAR and NAS-1Shot1 on the three different search spaces.}
\label{fig:nb101_ca}
\end{figure*}

Figure \ref{fig:nb101_ca} gives a comparison of the mean validation regret performance of 500 independent runs as a function of estimated training time forNAS-Bench-101 on the the three different CIFAR and NAS-1Shot1 on the three different search spaces.

Intuitively, we can observe the above six experimental results. Our experimental framework has the best effect on the six different search spaces contained in two different benchmark data sets,NAS-Bench-101\cite{Ying2019NAS-Bench-101} and NAS-Bench-1Shot1\cite{Zela2020aNAS-Bench-1Shot1}.
It can be seen that the algorithm based on the HB algorithm and the BOHB algorithm, compared with other comparison algorithms, the convergence speed is the fastest. This is because these algorithms try their best to find the optimal solution at the very beginning, compared to other algorithms. It further proves that Bayesian search strategy is much better than gradient search strategy.


For all three encodings in the search space, RS, TPE and RE, these three algorithms follow the same behavior at the beginning of the search. Finally, the evolutionary algorithms RE and DE produce better performance than the Bayesian search strategy algorithm. This further illustrates that predictor plays a deterministic role, while BOHB only plays an important role in the early stage of search. In the later stage of search, predictor plays a much greater role than BOHB.


\begin{figure*}[htb]
  \centering\small
  \begin{minipage}[t]{0.5\linewidth}
    \centering
    \includegraphics[width=\textwidth]{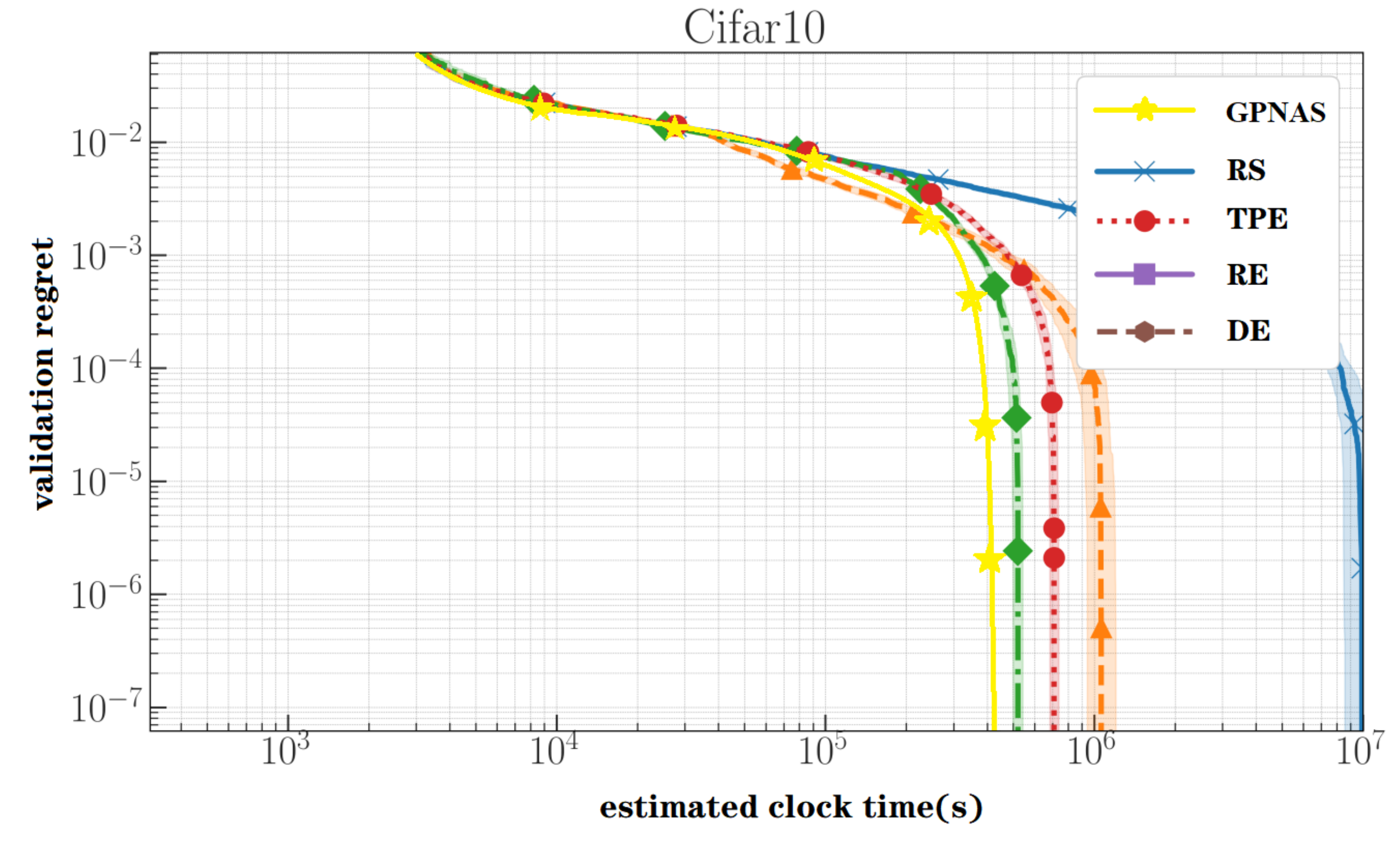}
   (a)  NAS-Bench-101 on CifarA
  \end{minipage}%
  \begin{minipage}[t]{0.5\textwidth}
    \centering
    \includegraphics[width=\textwidth]{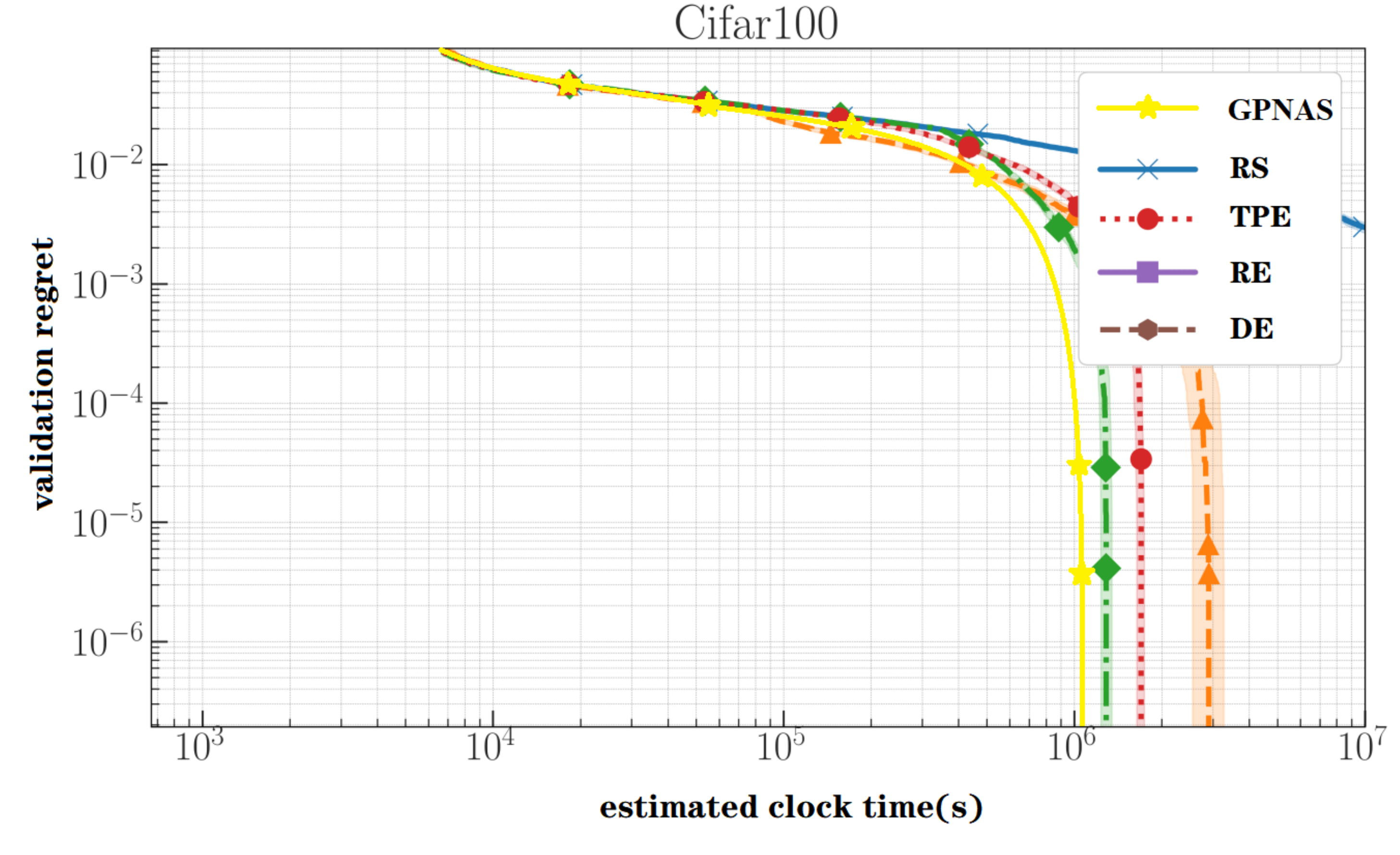}
    (b)  NAS-Bench-101 on  CifarB 
  \end{minipage}  \\
  \begin{minipage}[t]{0.5\textwidth}
    \centering
    \includegraphics[width=\textwidth]{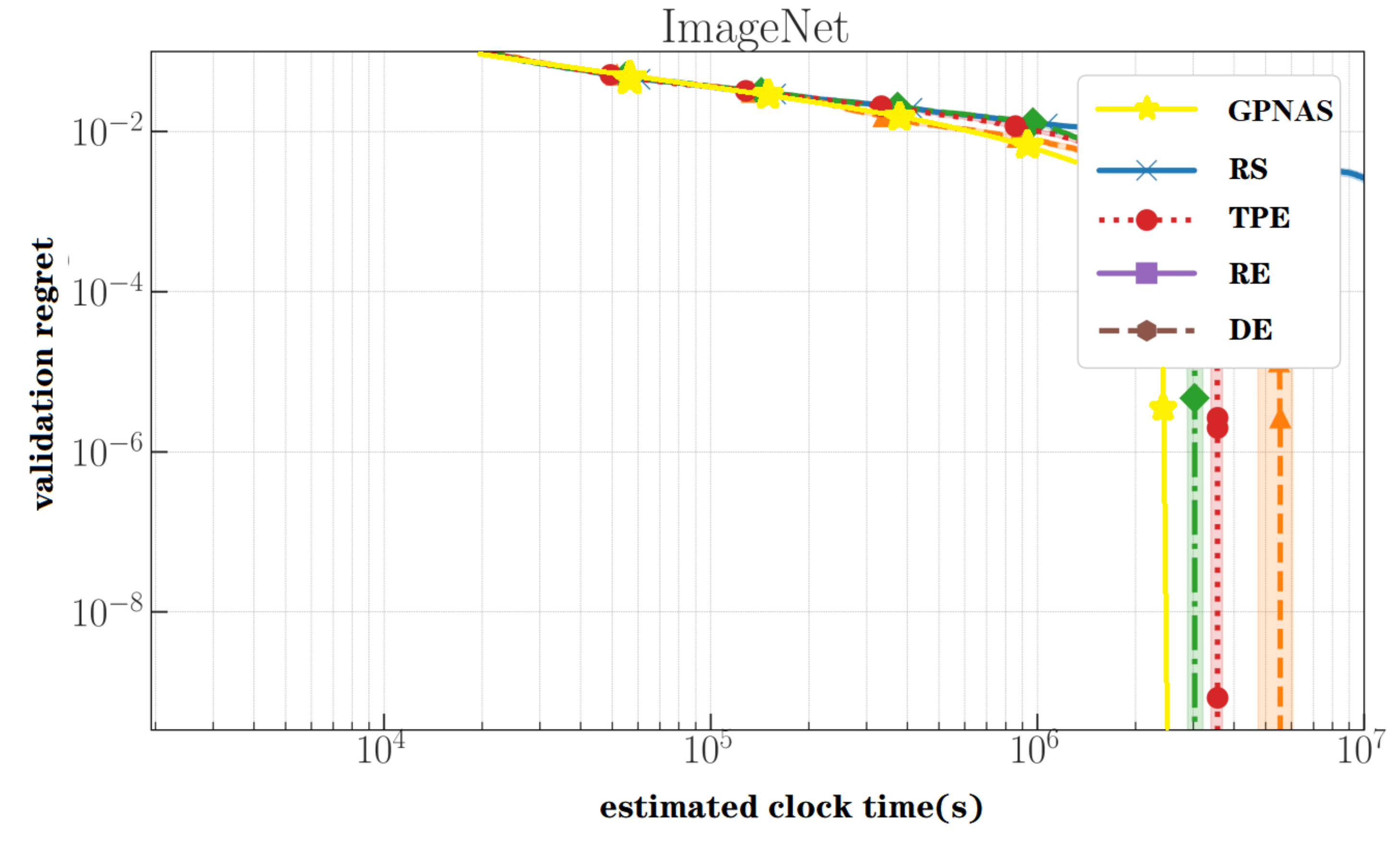}
    (c) NAS-Bench-101 on  CifarC
  \end{minipage}%
  \begin{minipage}[t]{0.5\textwidth}
    \centering
    \includegraphics[width=\textwidth]{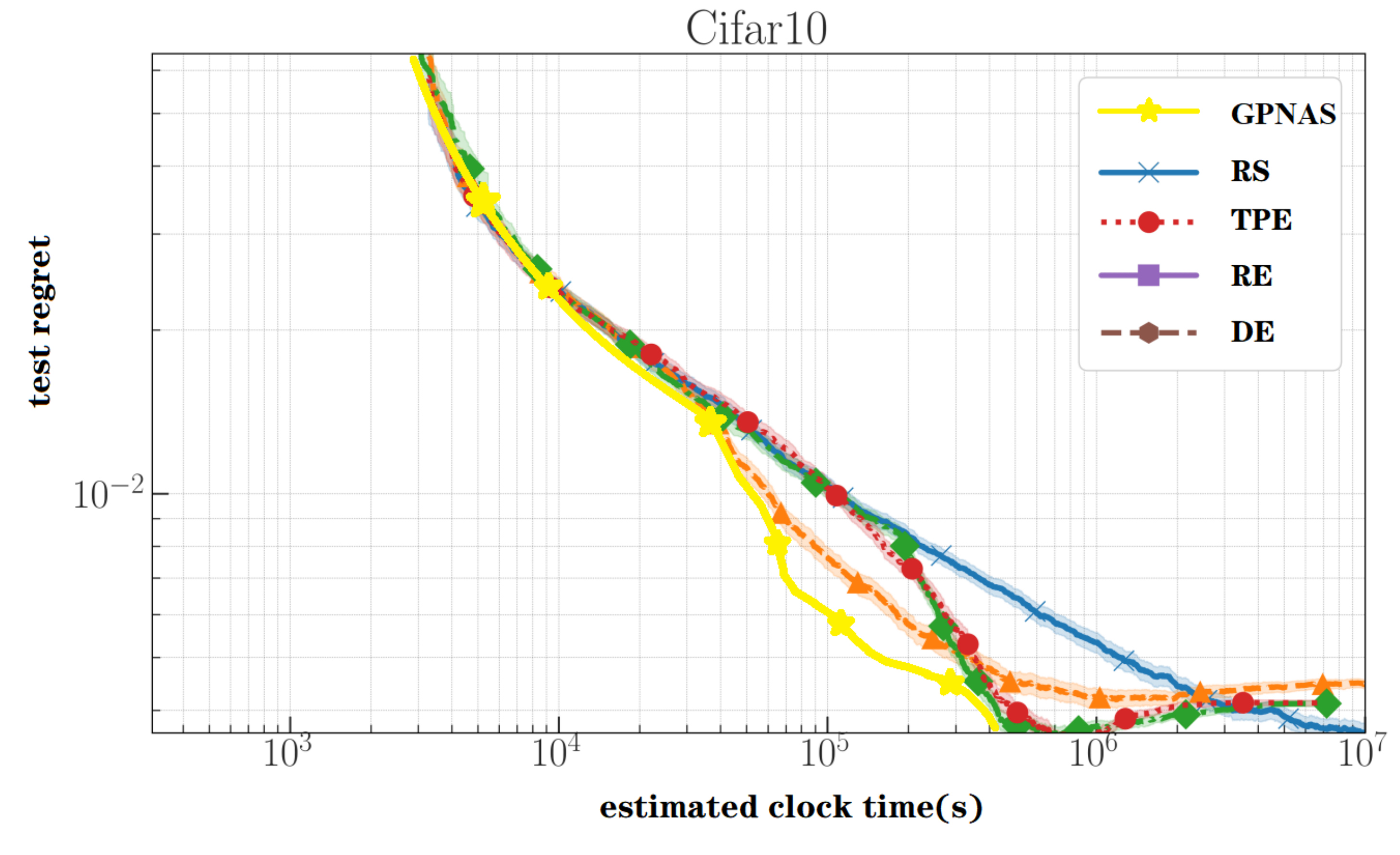}
    (d)  NAS-1Shot1 on the Search Space 1 
  \end{minipage}
    \begin{minipage}[t]{0.5\textwidth}
    \centering
    \includegraphics[width=\textwidth]{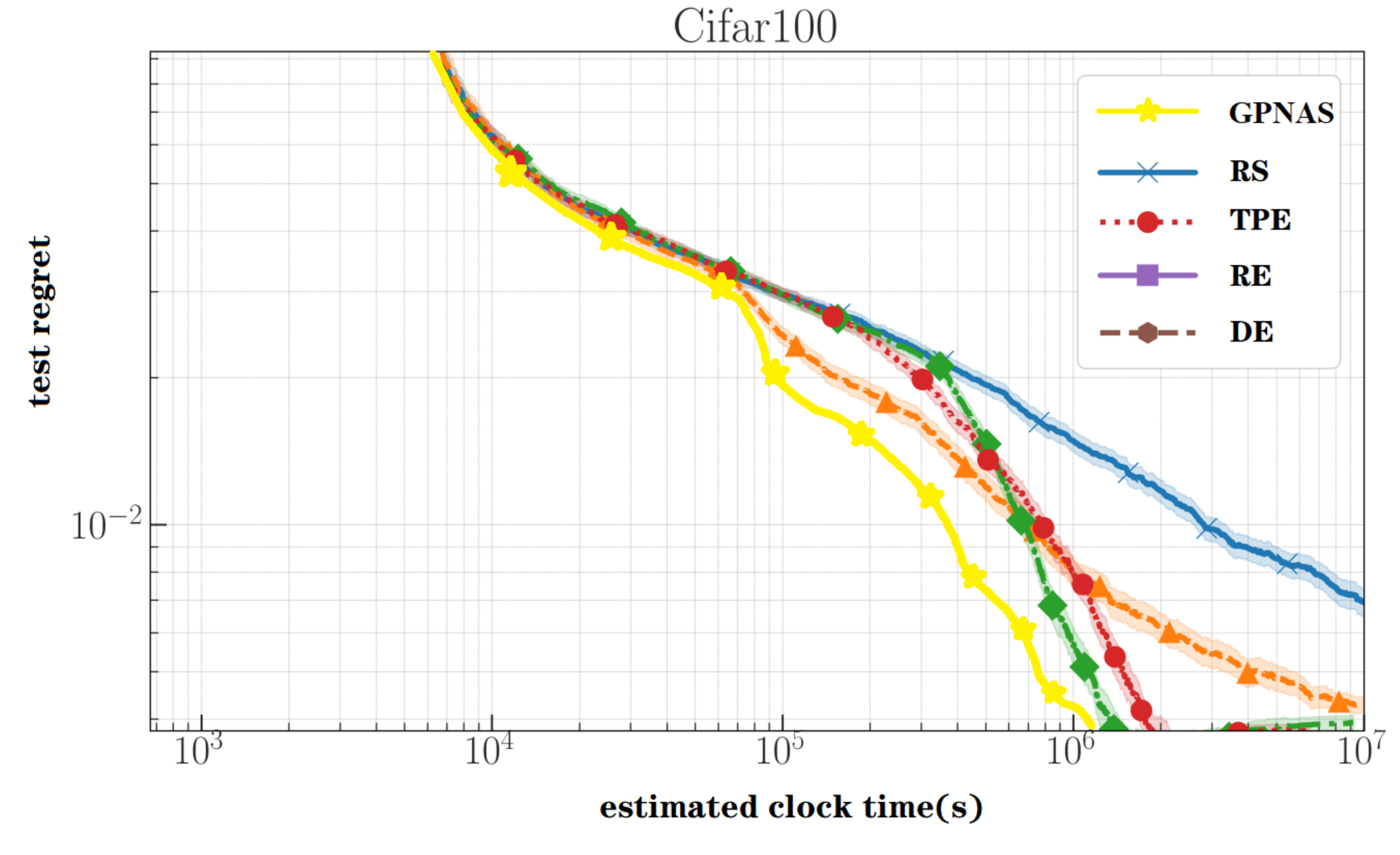}
    (e) NAS-1Shot1 on the Search Space 2 
  \end{minipage}%
  \begin{minipage}[t]{0.5\textwidth}
    \centering
    \includegraphics[width=\textwidth]{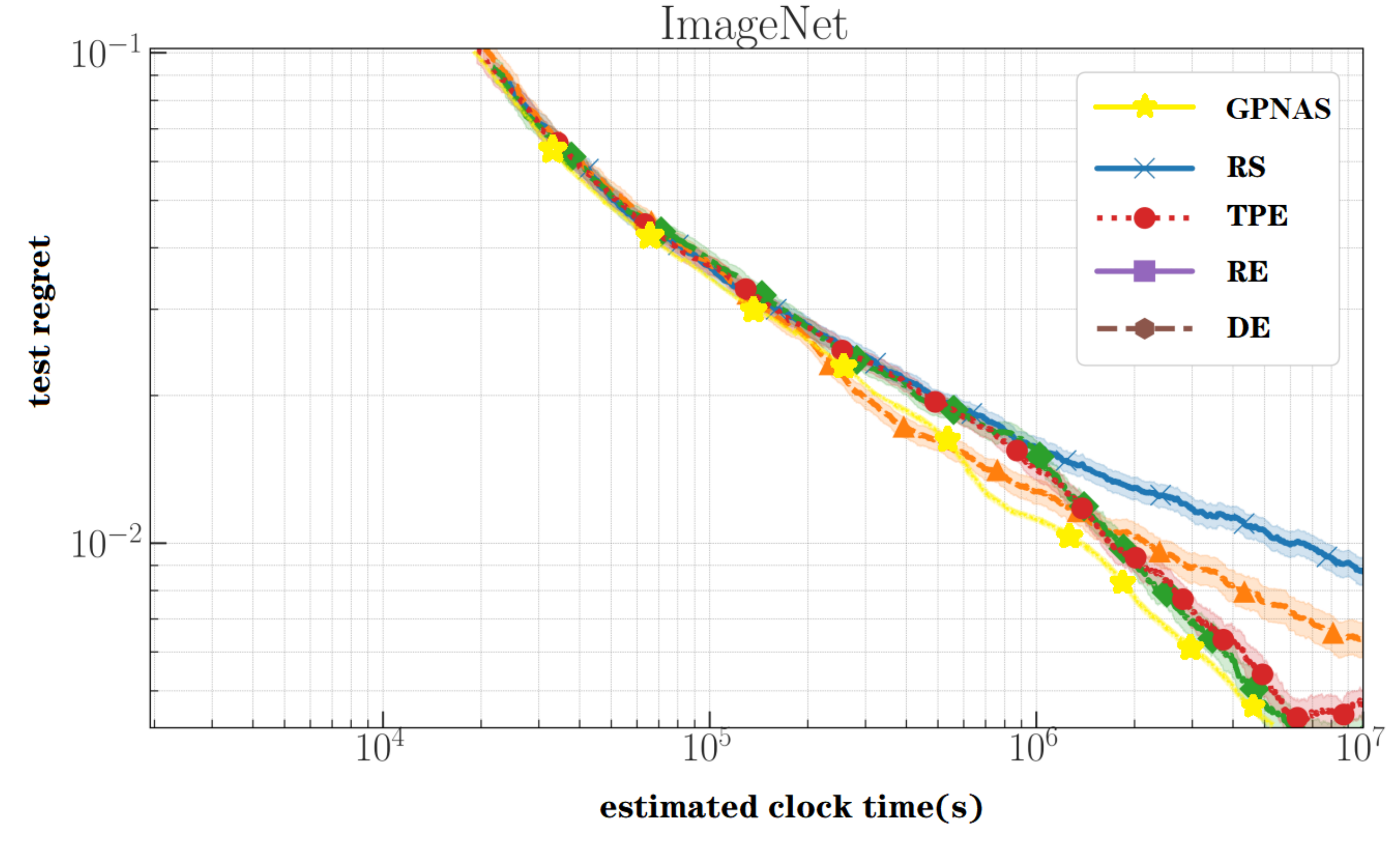}
    (f) NAS-1Shot1 on the  Search Space 3
  \end{minipage}
\caption{Sub-pictures (a)(b)(c) are comparison of the mean validation regret performance of 500 independent runs as a function of estimated training time for NAS-201 on Cifar10, Cifar100 and ImageNet and Sub-picture (d)(e)(f) is the comparison of the mean test regret performance of 500 independent runs as a function of estimated training time for NAS-201 on Cifar10, Cifar100 and ImageNet.}  
\label{fig:bessel-function}
\end{figure*}

In order to verify the generalization performance of our model. Next, we still use the experimental background of Awad\cite{Awad} to verify the generalization ability of the model on the three data sets (Cifar-10, CIFAR-100 and ImageNet16-120) to prevent our model from performing well on small data sets and not having the expected migration ability on large-scale data sets, even worse performance.

We evaluate five models (GPNAS, TPE, RS, RE, and DE) on three data sets (Cifar-10, CIFAR-100 and ImageNet16-120).
Figure \ref{fig:bessel-function} shows the results of the evaluation:
Sub-pictures (a)(b)(c) are comparison of the mean validation regret performance of 500 independent runs as a function of estimated training time for NAS-201 on Cifar10, Cifar100 and ImageNet,while sub-picture (d)(e)(f) is the comparison of the mean test regret performance of 500 independent runs as a function of estimated training time for NAS-201 on Cifar10, Cifar100 and ImageNet.

From the results, the following conclusions can be drawn. In sub-pictures (a), (b), and (c), the performance of validation regret is still impressive, whether it is on Cifar-10 or ImageNet16-120.

In sub-figure (d)(e)(f), the performance of test regret of the model shows that the robustness of all models is indeed reduced. The performance of TPE, RE and DE, the three models, is compared with GPNAS, the degree of fluctuation is more obvious, especially on the smaller the data set, the greater the degree of fluctuation. On the contrary, the larger the data set, the more gentle the trend of the performance curve. It can be roughly concluded that GPNAS, like TPE, RE and DE, can converge and converge faster.
\section{Conclusion}
In this paper, we design a better framework for NAS, using a dual-accelerator architecture search submodel. From the experimental conclusion, we achieved a very excellent effect under the constraints of resources. By comparing the Validation Regrets of multiple search methods on multiple search Spaces, our framework has a significant advantage over all non-gradient methods, thanks to the Predictor optimization. Benefiting from GCN, Predictor can assess the stability of a network architecture.Experiments show that some architectures may have higher stability when diversified operators are used. Most of the current work is to migrate cells to the NASNET architecture and evaluate cell performance in this architecture. There are reasons to believe that the NASNET architecture is not optimal, and that future work could use Predictor to discover new macroscopic architectures to replace the NASNET network architecture.
\bibliographystyle{splncs04}
\bibliography{main}
\vspace{12pt}
\color{red}

\end{document}